\documentclass[conference]{IEEEtran}
\usepackage{amsmath, amsfonts, amsthm, amssymb}
\usepackage{biblatex}
\usepackage{graphicx}
\usepackage{multirow}
\usepackage{hyperref}
\usepackage{caption}
\usepackage{flushend}
\newcommand{\oneto}[1]{1\!:\!{#1}}
\newcommand{\ind}[3]{({#1}\!\perp\kern-6pt\perp\!{#2}|{#3})}
\newcommand{\tot}[3]{t({#1}\rightsquigarrow {#2} || {#3})}

\newcommand{\E}{\mathbb{E}}
\newcommand{\R}{\mathbb{R}}

\newcommand{\BB}{\mathbf{B}}
\newcommand{\CC}{\mathbf{C}}
\newcommand{\EE}{\mathbf{E}}
\newcommand{\II}{\mathbf{I}}
\newcommand{\JJ}{\mathbf{J}}
\newcommand{\KK}{\mathbf{K}}

\newcommand{\TT}{\mathbf{T}}
\newcommand{\UU}{\mathbf{U}}
\newcommand{\XX}{\mathbf{X}}

\newcommand{\bb}{\mathbf{b}}
\newcommand{\cc}{\mathbf{c}}
\newcommand{\ee}{\mathbf{e}}
\newcommand{\bt}{\mathbf{t}}
\newcommand{\xx}{\mathbf{x}}

\newcommand{\SSigma}{\mathbf{\Sigma}}

\newcommand{\cE}{\mathcal{E}}
\newcommand{\cJ}{\mathcal{J}}
\newcommand{\cU}{\mathcal{U}}

\DeclareMathOperator{\avg}{avg}

\begin{document}
\title{Comparative Study of Causal Discovery Methods for Cyclic Models with Hidden Confounders}

\author{
\IEEEauthorblockN{Boris Lorbeer}
\IEEEauthorblockA{\textit{Technische Universit\"at Berlin} \\
Berlin, Germany \\
lorbeer@tu-berlin.de}
\and
\IEEEauthorblockN{Mustafa Mohsen}
\IEEEauthorblockA{\textit{Technische Universit\"at Berlin} \\
Berlin, Germany \\
m.mohsen@campus.tu-berlin.de}
}

\maketitle

\begin{abstract}
    Nowadays, the need for causal discovery is ubiquitous. A better
    understanding of not just the stochastic dependencies between parts of a
    system, but also the actual cause-effect relations, is essential for all
    parts of science. Thus, the need for reliable methods to detect causal
    directions is growing constantly.
    In the last 50 years, many causal discovery algorithms have emerged, but
    most of them are applicable only under the assumption that the systems have
    no feedback loops and that they are causally sufficient, i.e.\ that there
    are no unmeasured subsystems that can affect multiple measured variables.
    This is unfortunate since those restrictions can often not be presumed in
    practice. Feedback is an integral feature of many processes, and real-world
    systems are rarely completely isolated and fully measured. Fortunately, in
    recent years, several techniques, that can cope with cyclic, causally
    insufficient systems, have been developed. And with multiple methods
    available, a practical application of those algorithms now requires
    knowledge of the respective strengths and weaknesses.
    Here, we focus on the problem of causal discovery for sparse linear models
    which are allowed to have cycles and hidden confounders. We have prepared a
    comprehensive and thorough comparative study of four causal discovery
    techniques: two versions of the {\em LLC} method \cite{llc} and two
    variants of the {\em ASP}-based algorithm \cite{asp}. The evaluation
    investigates the performance of those techniques for various experiments
    with multiple interventional setups and different dataset sizes.
\end{abstract}


\section{Introduction}
Causal analysis \cite{pearl, pjs} of complex systems is nowadays an integral
part of many sciences. It is heavily used in fields as diverse as medicine,
biology, cognitive science, economics, predictive maintenance, root cause
analysis, physics, and machine learning.  Conventional data analysis
investigates the probabilistic properties of the data to gain insight into the
involved probability distributions which can then be used to e.g. predict new
data. Causality on the other hand serves to not just learn the data but to
learn about the system that generates the data. For instance, consider two
random variables, one is binary and indicates the administration of a drug that
is allegedly reducing blood pressure in patients, and the other is the
patient's blood pressure itself.
Then, ordinary data analysis can study the existence and size of a stochastic
dependency between those two random variables. But if we are interested in the
efficiency of the drug in changing blood pressure, this purely stochastic 
information is insufficient, since there can be a correlation between the two
random variables without causation, i.e. without the drug having any effect on
blood pressure. Causal analysis on the other hand provides techniques to answer
questions about actual causation. And in this example, it answers the question
of whether the drug is actually the reason for the change in the patient's blood
pressure and also measures the size of this causal effect. I.e. causal analysis is concerned
with the discovery and measurement of actual {\em mechanisms} in the underlying
system, which in this case would be the patient's body.

A subfield of causality is {\em causal discovery}, which studies the existence of causal
relations and is not concerned with the estimation of the size of the causal effects.
Often, causal discovery is the first step and its results serve as input for
the estimation of the effect size. This paper focuses on causal discovery.

Most of the research in causal analysis focuses on the acyclic situation, meaning
that the considered system is presumed to have no cycles in its causal
relations, i.e. there are no feedback loops. This renders the analysis less complex but
excludes many realistic scenarios. An example from econometrics would be the study
of supply, price, and demand: the demand is influencing the price but the price is also
influencing the demand.

Another assumption in standard causal analysis is {\em causal sufficiency}: It is
presumed that there are no unobserved variables, so-called {\em hidden confounders},
that causally influence multiple observed variables. Again, this simplifies the 
analysis but excludes many relevant use cases. E.g., in the example above of a 
drug for blood pressure, imagine that the drug is only given to younger people,
which have lower blood pressure anyways, thus causing bias to the results.
In this case, age is a hidden confounder.

While there are many causal discovery algorithms for the acyclic, causally
sufficient situation, very few exist that are also applicable in the more
demanding case of cyclic systems with hidden confounders. In this paper,
we compare the properties of four of such methods, namely two techniques,
{\em ASP-d} \cite{asp} and {\em ASP-s} \cite{sigmasep},
which are variants of a constraint-based technique using answer set programming,
and two variants, {\em LLC-NF} and {\em LLC-F} \cite{llc}, of a method of
moments type estimator.

Those four approaches are evaluated on synthetic data from linear systems,
which means the causal relationships between the variables are linear.

\section{Related Work}
For a complete overview of the history of causality see the treatises
\cite{pearl} and \cite{pjs}. Both references focus mainly on the acyclic case
but do cover, to a certain extent, the situation with hidden confounders.

An early technique for cyclic systems, called {\em CCD}, is described in
\cite{richardson2013discovery}, but it presumes the absence of latent
confounders. Amongst the few algorithms that allow for both cycles and hidden
confounders are {\em LLC} \cite{llc}, the method described in \cite{asp} which
we refer to as {\em ASP-d}, {\em sigmasep} \cite{sigmasep} which we refer to as
{\em ASP-s}, {\em BACKSHIFT} \cite{rothenhausler2015backshift}, {\em CCI}
\cite{strobl2019constraint}, and {\em bcause} \cite{bcause, sBcause}.

Note that the methods above presume interventional data, that is, not only data
from the system itself but also from other systems that are obtained by
changing the original system in a certain way, i.e. {\em intervening} on it. If
this interventional data is not available, causal inference becomes harder.
There are, however, several approaches for this scenario, too, like the family
of {\em Additive Noise Models} ({\em ANM}), see e.g. \cite{hoyer2008nonlinear},
ICA-based methods for linear systems like {\em LiNGAM} \cite{lingam}, {\em
LiNG} \cite{lacerda2012discovering}, and the {\em Two-Step} algorithm
\cite{twostep}, or {\em Information Geometric Causal Inference} ({\em IGCI}),
see e.g.  \cite{mooij2016distinguishing}. Some of those methods can also deal
with latent confounders or cyclicity.

In recent years, the mathematical foundations of the theory of cyclic systems with
hidden confounders have advanced considerably. A comprehensive exposition can be found
in \cite{forre2017markov} and \cite{bongers}.

\section{Description of the Methods}\label{secMethods}
This section will present the main features of the {\em LLC} and {\em ASP} algorithms.
Note, that this is a high-level overview, presenting only as much as is
necessary to explain the evaluation below. For the details, the reader should
consult the pertinent papers, see \cite{eberhardt2010combining, llc-f, llc} for
the {\em LLC} and \cite{asp, sigmasep} for the {\em ASP} variants.

\subsection{General Concepts in Causality}
The main entity in causality is the {\em Structural Causal
Model} (SCM) \cite{bongers}, which consists of
{\em structural equations} of the form (note that we use $\oneto n$ as an abbreviation
for the set $\{1,2,\ldots,n\}$):
\begin{equation}
X_i = f_i(\XX, \EE), \qquad i\in \oneto n,
\end{equation}
where $\XX = (X_1,\ldots, X_n)$ is an $n$-dimensional random vector containing
the observed variables, $\EE$ is an $m$-dimensional random vector containing
the unobserved noise, and the functions $f_i:\R^n\times\R^m\to\R$ are the {\em
causal mechanisms}. See \cite{bongers} for the mathematical details. Many
aspects of an SCM can be described by associating a graph
that contains as nodes both the observed variables $X_i, i\in\oneto n$ and the
unobserved variables $E_k, k\in\oneto m$. It is a {\em directed graph} (DG), i.e.
all edges have exactly one arrowhead.
Here, a directed edge only goes into observed nodes $X_i$, no edges are pointing to
noise variables $E_k$, and there is an edge $N\to X_i$ from a node $N$ to $X_i$
iff $f_i$ depends on $N$. This graph is called the {\em augmented graph of the
SCM}, see \cite{bongers}.

This directed graph can be reduced to a graph that contains as nodes only the
observed variables and connects any two nodes that have a noise node as common
parent by a bidirected edge. Graphs with both directed and bidirected edges are
called {\em directed mixed graphs} (DMG). This reduced graph is simply called
the {\em graph of the SCM}.

A {\em path} (in a DG or DMG) is a tuple $(\epsilon_1,\ldots,\epsilon_m)$ of
edges where two consecutive edges have a common node. In particular, an edge
can appear multiple times in a path. A path is a {\em directed path} if none of
the edges are bidirected and all edges point in the same direction.

The directed edges in the graph of an SCM then depict the direct causal
connections between variables, symbolizing {\em direct causal effects}, while
bidirected edges describe confounding (see below).
Directed paths consisting of more than one edge describe indirect causal
connections, generating {\em indirect causal effects}. Note, however, that in
cyclic SCMs we can have causal effects even between nodes that are not
connected by a directed graph, see \cite{bongers} section 7.1.

If this graph has {\em cycles}, i.e. directed paths that start and end at the same
node and contain at least one edge, the SCM is called {\em cyclic}. Causal cycles describe
systems with feedback loops, which are quite common in realistic scenarios.

If an unobserved node $E_k$ influences two different observed nodes $X_i, X_j$,
i.e. the augmented graph of the SCM contains the subgraph $X_i\leftarrow E_k
\to X_j$ and the graph of the SCM contains a bidirected edge
$X_i\leftrightarrow X_j$, then $E_k$ is called a {\em hidden} (or {\em latent})
{\em confounder}. A system without hidden confounders is called {\em causally
sufficient}.

A central notion in causality is that of an {\em intervention}. We consider only
{\em surgical interventions} \cite{llc}, which can be described as follows: The original
SCM is changed by selecting a subset $\XX_I, I\subset \oneto n$, of the observed
variables, and forcing a new distribution on $\XX_I$, that is independent of
all the other random variables $\XX_{1:n\setminus I}$ and $\EE$. An example would be
randomized drug trials, which ensure that the people who do and do not get the
drug are selected completely at random. In this case, we have an intervention
on the assignment of the drug. The graph of the intervened SCM differs from the
graph of the original SCM in that all the edges that point into intervened
nodes are removed.

Data that is observed from a non-intervened system is called {\em (purely)
observational data}.

A common approach to causal discovery consists in using {\em constraint-based}
methods, which exploit conditional (in)dependences between random variables to
infer causal connections. Two random variables $X$ and $Y$ are said to be
independent conditioned on a set of random variables $S$ with $X,Y\notin S$,
denoted by $\ind{X}{Y}{S}$, if they are independent w.r.t. their conditional
probabilities, i.e.:
\begin{equation}
    \ind{X}{Y}{S}\quad \Leftrightarrow \quad p(X, Y | S) = p(X|S) p(Y | S),
\end{equation}
presuming those conditional probabilities exist. The notion of ``conditional
independence'' has a pendant in graphs which is called {\em $d$-separation}. To properly
explain it, we first need to define colliders: Given
a path in a DMG, a node $X_i$ on the path is a {\em collider on this path}
if both neighboring edges on the path point into $X_i$, i.e.
${X_{i-1}\to X_i\leftarrow X_{i+1}}$,
${X_{i-1}\leftrightarrow X_i\leftarrow X_{i+1}}$,
${X_{i-1}\to X_i\leftrightarrow X_{i+1}}$, or
${X_{i-1}\leftrightarrow X_i\leftrightarrow X_{i+1}}$.
Then two nodes $X_i$ and $X_j$ are said to be {\em d-connected}
w.r.t.\ a conditioning set $C$ if there is a path from $X_i$ to $X_j$ such that
all the colliders on the path are in $C$. If two nodes are not d-connected they
are called {\em d-separated}. The notation for $X$ being d-separated from $Y$
given $C$ in the graph G is ${(X\perp_G Y | C)}$.

The idea of constraint-based methods is based on two assumptions (see \cite{pearl,
pjs} for details):
\begin{enumerate}
    \item The probability distribution of an SCM is {\em Markovian} w.r.t.\
        the graph $G$ of the SCM, i.e.:
        \begin{equation}
            (X\perp_G Y | C) \qquad \Rightarrow \qquad \ind{X}{Y}{C}.
        \end{equation}
        For linear, possibly cyclic SCMs, which is the case we are considering,
        this assumption holds under mild conditions. See Theorem A.7 in
        \cite{forre2017markov} for a precise statement of those conditions.
    \item The probability distribution of an SCM is {\em faithful} w.r.t.\
        the causal graph $G$ of the SCM, i.e.:
        \begin{equation}
            \ind{X}{Y}{C} \qquad \Rightarrow \qquad (X\perp_G Y | C).
        \end{equation}
        There are situations when this implication does not hold. E.g., imagine
        for an edge $X\to Y$ a second path $X\to Z\to Y$ which creates an
        effect from $X$ on $Y$ that is exactly the opposite of that of the edge
        $X\to Y$. Thus, the two effects cancel out, and, even though the system
        has a proper mechanism that links $X$ and $Y$, it is invisible in the
        data.
\end{enumerate}
Now, constraint-based methods usually presume those two assumptions and use
them to obtain information about the causal graph structure from stochastic
(in)dependence properties of the observational and interventional data. The
advantage of the constraint-based approach is that it is nonparametric, i.e. we
don't have to presume a specific model for the SCM. Note, however, that it only
provides the structure of the graph of the SCM. For the quantitative estimation
of the causal effects, one has to use further techniques on top of the
constraint-based methods.

Note, that with constraint-based methods one tries to obtain the non-symmetric
property ``$X$ causes $Y$'' from symmetric stochastic (in)dependence properties
which is usually not possible. Thus, one has to apply interventions: If $X$
causes $Y$, i.e. $X\to Y$, then both $Y$ is stochastically dependent on $X$ and
$X$ is stochastically dependent on $Y$.  But with intervention on $X$, the
dependence still holds, while intervening on $Y$ removes the dependence, thus
breaking the symmetry.

\subsection{The {\em LLC} algorithm}\label{secLLC}
This section gives a high-level overview of the {\em LLC} algorithm; for more
details, see \cite{llc}.

{\em LLC} is a method of estimating the parameters of a {\em linear} causal system of the form:
\begin{equation}\label{base}
    \xx = \BB\xx + \ee
\end{equation}
where $\xx,\ee\in\R^n, \BB\in\R^{n\times n}$. Here, $\xx$ contains the
measurements of the observed variables $\XX$ and $\ee$ the hidden values of the
unobserved variables $\EE$. For $\EE$, we only presume the expectation to be zero,
$\E[\EE] = 0$, and the covariance of $\EE$ is abbreviated with $\SSigma_\ee$. Note,
that we do not require any particular distribution for $\EE$, the only restriction
is $\E[\EE] = 0$, and even that can be lifted, see \cite{llc}. However, since this
change has no bearing on our evaluation, we stick to the simple case $\E[\EE] = 0$.
The $(i,j)$-coefficient of the matrix $\BB$ is denoted by $b_{ij}$ and can be
identified as the direct causal effect of variable $X_j$ on variable $X_i$.
$\BB$ is allowed to describe cyclic paths in the causal graph.  The covariance
$\SSigma_\ee$ is allowed to have off-diagonal elements, which can be
interpreted as confounding. $\BB$ is required to have a zero diagonal:
\begin{equation}
    b_{ii} = 0,\qquad i=1,\ldots,n,
\end{equation}
which translates to the system not having self-loops.

The measurement of an intervened system, possibly with an empty intervention,
is called an {\em experiment}. Experiments are
denoted by $\cE_k := (\cJ_k, \cU_k), k\in \oneto K$, where $K$ is the number
of considered experiments and each $(\cJ_k,
\cU_k)$ is a partition of $\oneto n$, i.e. $\cJ_k\cap \cU_k=\emptyset$ and
$\cJ_k\cup\cU_k=\oneto n$. Here, $\cJ_k$ is the set of indices of the nodes which
are intervened on and $\cU_k$ of those that are not. To each experiment $\cE_k=(\cJ_k, \cU_k)$
a pair of diagonal matrices $(\JJ_k, \UU_k)$ is assigned: using the indicator function $I(\cdot)$,
which equals one if its argument is true and zero otherwise,
the diagonals of $\JJ_k$ and $\UU_k$ are given by $\JJ_{k,ii} = I(i\in\cJ_k)$ and
$\UU_{k,ii} = I(i\in\cU_k)$, resp.
This allows us to write the structural equations for the experiment $\cE_k$ in a
very compact form. Let $\cc$ be the vector containing at the indices $\cJ_k$ the
intervention values (the values the pertaining variables are forced to), and
zeros elsewhere. Then the structural equations for the experiment $\cE_k$ are given
by:
\begin{equation}\label{baseInt}
    \xx = \UU_k\BB\xx + \UU_k\ee + \cc.
\end{equation}

For {\em LLC} to work, it is required that the SCM be {\em weakly stable}, i.e.
for every experiment $\cE_k = (\cJ_k, \cU_k)$, the matrix $\II - \UU_k\BB$ must
be invertible.

Next, we consider the covariance matrix $\CC_\xx^k$ of the measurements $\XX$ in
experiment $\cE_k$. A covariance $c_{ui}$ in $\CC_\xx^k$ is called the {\em total
causal effect} of $x_i$ on $x_u$ with intervention set $\cJ_k$, and is abbreviated
as $t(x_i\rightsquigarrow x_u || \cJ_k)$. A central identity of {\em LLC} provides
the connection between those total causal effects and the matrix $\BB$, i.e. the
direct causal effects. This identity reads:
\begin{equation}\label{total}
    t(x_i\rightsquigarrow x_u || \cJ_k) =
    b_{ui} + \sum_{j\in\cU_k\setminus\{u\}} t(x_i\rightsquigarrow x_j || \cJ_k)b_{uj},
\end{equation}
which is a linear equation in the $b_{ij}$. Gathering all those equations for all
conducted experiments results in a system of linear equations:
\begin{equation}\label{tTb}
    \bt = \TT \bb,
\end{equation}
where $\bt$ contains all the total effects on the left-hand side of (\ref{total}),
$\bb$ is the concatenation of all the rows of $\BB$ without the diagonal, i.e.
$\bb\in\R^{n^2-n}$, and $\TT$
contains the total effects from the right-hand side of (\ref{total}). 
In \cite{llc} it is shown, that for $\TT$ in (\ref{tTb}) to have
zero null space, the experiments $\{\cE_k\}_{k=1}^K$
need to satisfy the {\em pair condition}, which requires that for each ordered pair of
indices $(i, j), i,j\in\oneto n$, there is an experiment $\cE=(\cJ, \cU)$ such that
$i\in\cJ, j\in\cU$.

Thus, the method of {\em LLC} becomes clear: First, the covariances $\CC_\xx^k$ are
estimated from the data, those total effects are used to build the linear
system (\ref{tTb}), and finally, this linear system is solved for $\bb$.

But {\em LLC} also provides an estimate for the covariance matrix $\SSigma_\ee$, which
describes the confounding in the system.
If purely observational (i.e. non-intervened) data is available, equation
(\ref{base}) applies and since we obtained an estimate for $\BB$ from
(\ref{tTb}), we can simply compute $\SSigma_\ee$ as:
\begin{equation}\label{cofVar}
    \SSigma_\ee = (\II-\BB)\CC_\xx^0(\II - \BB)^T,
\end{equation}
where $\CC_\xx^0$ is the covariance matrix of $\xx$ for no intervention.
If there is no purely observational data, we can still obtain $\SSigma_\ee$ from
the experiments: from (\ref{baseInt}) it follows for any experiment $\cE_k=(\cJ_k, \cU_k)$ that:
\begin{equation}
    (\SSigma_\ee)_{\cU_k, \cU_k} = \left[(\II - \UU_k\BB)\CC_\xx^k(\II-\UU_k\BB)^T\right]_{\cU_k, \cU_k}.
\end{equation}
Depending on the experiments, for a given pair $(i,j)$ there can be multiple experiments
$\cE_k=(\cJ_k,\cU_k)$ with $i,j\in\cU_k$, so it makes sense to compute the average
of the covariances over all such experiments, i.e.:
\begin{equation}
    \SSigma_{\ee,ij} =
    \avg
        \left\{
            \left[(\II - \UU_k\BB)\CC_\xx^k(\II-\UU_k\BB)^T\right]_{i,j}
            |i,j\in\cU_k
        \right\}.
\end{equation}
Thus, to obtain the complete covariance $\SSigma_\ee$, we need for each pair $(i,j)$
at least one experiment $\cE_k = (\cJ_k, \cU_k)$ such that $i,j\in\cU_k$, which
is called the {\em covariance condition}.

Finally, if all conditions above are satisfied, {\em LLC} returns the pair $(\BB,
\SSigma_\ee)$.
Note, that in causal discovery, we are only interested in the causal graph,
i.e.\ we only need to know which $b_{ij}$ and $\SSigma_{\ee,ij}$ are zero.

\subsection{The {\em LLC-F} algorithm}
Note, that {\em LLC} is not a constraint-based method, and does not presume
faithfulness. However, the question is whether combining {\em LLC} with
constraint-based methods could improve the accuracy of {\em LLC.} This has been
investigated in \cite{llc-f} and the pertinent algorithm is called {\em LLC-F,} with the
additional letter ``F'' indicating that now, faithfulness is presumed. We use the
abbreviation {\em LLC-NF} to refer to the {\em LLC} variant that does not use constraint-based
methods and does not presume faithfulness.

The idea is to add to the linear system (\ref{tTb}) more linear equations obtained
from conditional independences in the purely observed and intervention data. The
following four methods are applied:
\begin{enumerate}
    \item If, for some experiment $\cE_k = (\cJ_k, \cU_k)$, for two
        variables $X_i, X_j$ with $i,j\in\cU_k$ there exists a set $S$ of
        variables with $X_i,X_j\notin S$ such that $\ind{X_i}{X_j}{S}$, then we
        have, by faithfulness, $b_{ij} = b_{ji} = \SSigma_{\ee,ij} = 0$.
    \item If, for some experiment $\cE_k = (\cJ_k, \cU_k)$ and two variables $X_i,
        i\in\cJ_k$ and $X_u, u\in\cU_k$, there exists a set $S$ of variables with
        $X_i,X_u\notin S$ such that $\ind{X_i}{X_u}{S}$, then we have, by
        faithfulness, $b_{ui} = 0$.
    \item If, for some experiment $\cE_k = (\cJ_k, \cU_k)$ and three indices
        $i\in\cJ_k$ and $u,v\in\cU_k$, we have $\tot{x_i}{x_u}{\cJ_k}\ne 0$ and
        $\tot{x_i}{x_v}{\cJ_k} = 0$, then it follows that $b_{vu}=0$ by
        faithfulness.
    \item If, for some experiment $\cE_k = (\cJ_k, \cU_k)$ and three indices
        $i\in\cJ_k$ and $u,v\in\cU_k$, we have $\tot{x_i}{x_u}{\cJ_k}\ne 0$ and
        $\ind{x_i}{x_v}{\{x_u\}}$ in $\cE_k$, then it follows that $b_{uv}=0$
        and $\SSigma_{\ee,uv} = 0$.
\end{enumerate}
Apart from extending the linear system (\ref{tTb}) with those equations, the
algorithm does not differ from {\em LLC-NF}.

\subsection{{\em ASP-d}}\label{secAspd}
Next, we give an overview of the {\em ASP-d} algorithm, following \cite{asp}. {\em ASP} is
an abbreviation of ``Answer Set Programming'', which is a declarative
programming language, see \cite{aspLogic}.

The {\em ASP-d} algorithm belongs to the class of constraint-based methods and
thus infers the causal graph from conditional independences as has been
described above. It differs from most other constraint-based techniques in that
it allows for cyclic causal graphs with hidden confounders and that it can deal
with data from multiple experiments with different interventions. The conditional independences are
obtained from independence hypothesis tests which have a certain probability of
error. That means they can contradict each other. The innovation of {\em ASP-d}
is to handle this issue by formulating this causal discovery problem as an
optimization problem: let $\KK$ be a set of conditional (in)dependence
relations that are obtained from a given dataset, and let $w(k)\in\R_{\ge 0}$
be a nonnegative weight assigned to each $k\in \KK$, describing how confident
we are that $k$ is indeed true. Then the task is to find a graph $G^\ast$ which
minimizes the following loss function:
\begin{equation}\label{loss}
    L(G) := \sum_{k\in\KK, G\nvDash k} w(k),
\end{equation}
where $G\nvDash k$ means that the graph does not entail, via $d$-separation,
the (in)dependence $k$. Since the loss function is using $d$-separation, {\em ASP-d}
is applicable to {\em acyclic} SCMs that are linear or nonlinear, and to {\em
cyclic} SCMs that are linear; see the description of {\em ASP-s} below for more
details.

Several possibilities for how to determine the weights $w(k)$ have been proposed, see
e.g.\ \cite{asp, sigmasep, sBcause}. We will presume that the constraints
$k\in\KK$ have been obtained using conditional independence hypothesis tests with
significance level $\alpha$ and then use weights as in \cite{sigmasep}:
\begin{equation}
    w(k) = |\log p_k - \log\alpha|,
\end{equation}
where $p_k$ is the p-value of the hypothesis test of $k$.

The optimization of (\ref{loss}) is done by encoding the problem as an {\em ASP}
program which can then be solved by some {\em ASP} solver like e.g. {\tt clingo}
\cite{clingo}.

\subsection{{\em ASP-s}}
In constraint-based methods, the (in)dependences obtained from the measurements
must be matched with the graph of the SCM to discover its causal structure.
This matching is done, as explained above, via $d$-separation. However, it should
be noted that for {\em nonlinear} cyclic SCMs, $d$-separation in general fails
to be Markovian, see \cite{forre2017markov}.

I.e., for nonlinear cyclic models, $d$-separation must be replaced with another
separation property. This new separation property {\em $\sigma$-separation} has
been introduced in \cite{forre2017markov}. The basic idea here is roughly that
nodes in loops are so strongly connected that conditioning cannot separate
them, so they behave as if they would be fully connected. That means, if loops
are replaced by fully connected subgraphs, an operation which is called an {\em
acyclification}, see \cite{forre2017markov, bongers} for details, the resulting
acyclic graph exposes $d$-separation properties that again correspond to the
conditional (in)dependences of the original cyclic SCM. In
\cite{forre2017markov} the authors then formulated a separation property for
cyclic graphs, {\em $\sigma$-separation}, which is the ``pull back'' of
$d$-separation via the acyclification operation.

Then, the authors of \cite{sigmasep} took the {\em ASP-d} algorithm and changed the
{\em ASP} encoding slightly to now use $\sigma$-separation instead of $d$-separation.
In \cite{sBcause} this algorithm got further improved to the {\em ASP-s}
algorithm we use in this paper.

Thus, {\em ASP-s} differs from {\em ASP-d} only in the type of separation that is used. In
particular, {\em ASP-s,} too, is a constraint-based, nonparametric, optimization
algorithm that minimizes (\ref{loss}), except that for {\em ASP-s} the notation
$G\nvDash k$ under the sum now refers to $\sigma$-separation.

It is important to note, however, that {\em ASP-d} and {\em ASP-s} have different
application fields. While {\em ASP-d} can be used for acyclic and {\em linear} cyclic SCMs
with hidden confounders, see above, {\em ASP-s} applies to acyclic and {\em nonlinear}
cyclic SCMs with hidden confounders. In particular, linear cyclic SCMs are not
faithful w.r.t.\ $\sigma$-separation, see \cite{forre2017markov, sigmasep}.
Thus, those two application fields have only the acyclic SCMs in common. And
since this paper only examines linear SCMs, {\em ASP-s} is not strictly applicable
here.  However, we included {\em ASP-s} in the set of algorithms to compare. Since
linear models are ``of measure zero'' inside the set of all SCMs, one might be
tempted to always use $\sigma$-separation. This could be problematic, because
nonlinear SCMs that are nearly linear could provide data that is only bearly
distinguishable from that of linear SCMs. Thus, it would be instructive to see
how much worse {\em ASP-s} performs compared to {\em ASP-d} on linear cyclic models.

\section{Evaluation}
Here we describe our evaluation of the four methods {\em LLC-NF}, {\em LLC-F},
{\em ASP-d}, and {\em ASP-s}. The source code of our implementation in {\em Python}
and {\em R} is available
online\footnote{https://github.com/mustafa-mohsen/llcsig}, and uses publicly
available code for {\em LLC} from Antti
Hyttinen's homepage\footnote{https://www.cs.helsinki.fi/u/ajhyttin/} and for
{\em ASP} from the GitHub repository of
\cite{sigmasep}\footnote{https://github.com/caus-am/sigmasep}.

For various types of data, we measure the methods' capabilities of detecting
features of the underlying SCM. Those features are edges and bidirected edges
(confounders), which can be either absent or present. The evaluation thus
consists of measuring the performance of binary predictions for each single
feature. Recall that there are requirements on the experimental setups for the
SCMs to be theoretically identifiable. Below, we consider the performances of
the methods for both cases, where the experimental setups do and do not satisfy
those requirements.

As described in Section \ref{secLLC}, the {\em LLC} variants are based on solving the
system (\ref{tTb}). For real data this system
must be expected to contain contradicting equations. This could be handled by
solving it as a least squares problem. However, as stated above, the system
could also have a non-zero null space if there are not sufficiently diverse
experiments to satisfy the pair condition. This would result in the indeterminacy
of some or all of the coefficients in $\bb$. To avoid this, we chose to use the
version of {\em LLC} that applies a penalty term ($L_1$ or $L_2$) to
(\ref{tTb}), see the code base of \cite{llc}, which also has the useful effect
of regularization and promoting sparsity.

The covariance matrix is fully determined if the covariance condition is
satisfied, see Section \ref{secLLC}. This is ensured by adding the
``null-experiment'', i.e.\ the experiment without any intervention, to each
of our experimental setups.

The {\em ASP}-based methods, being constraint-based methods, might not be able to
determine the presence of some features if there are not sufficiently many
interventions available. One possible approach in this scenario is to fix the
presence of undetermined features according to domain knowledge. For example,
if it is known that the system under consideration corresponds to a sparse
causal graph, a straightforward practice is declaring those undetermined
features as being absent. This can be considered as an ensemble of two methods,
the {\em ASP} algorithm and a weak classifier which classifies everything as absent.
The ensemble consists in applying first {\em ASP} and returning its result unless it
is undetermined in which case the weak classifier is applied. Most real-world
causal graphs are sparse, so it is reasonable to confine our evaluations to
data from sparse causal graphs and to always use the above ensemble. Thus,
below, whenever we refer to {\em ASP-s} or {\em ASP-d,} we actually refer to this
ensemble.

Since there are almost no real-world datasets with known causal ground truth
available, let alone in sufficient quantities and satisfying the particular
constraints required by our evaluation, we restrict our study to synthetic
data.
Because of the high computational complexity of the methods, in particular of
the {\em ASP} variants, we simulate only graphs with five nodes and two
confounders. The edges are randomly distributed with the constraint that the
in-degree of any node is not larger than two. As a result, the average number
of edges in our simulated graphs is 6.1 and that of bidirected edges is two,
i.e.\ the simulated graphs are sparse.

The coefficients of the linear equations in the SCM are sampled from the uniform
distribution over the set $[-1.1,-0.1]\cup[-0.1,1.1]$. The effect sizes are
thus bounded away from zero, assuring detectability. Furthermore, It is
ascertained that each simulated SCM has at least one cycle (which is not a
self-loop). The study in this paper is based on data generated by 150 such
randomly sampled SCMs.

For the evaluation of the methods, we need a metric. Both the {\em
LLC} and the {\em ASP} algorithms return for each feature a score determining
how strongly the algorithm ``believes'' this feature to be present. The code
from \cite{llc} also provides a bootstrapped version of {\em LLC} which means
we obtain for each coefficient $b_{ij}$ and $\SSigma_\ee$ a collection of
estimates of which we can compute the z-score. This is the score we use for
both {\em LLC} algorithms. For the {\em ASP} variants, we utilize a score proposed
in \cite{magliacane2016ancestral}: the {\em ASP} algorithm is run twice for
every single feature, once with the additional constraint that the feature is
present and once with the additional constraint that it is absent. The score is
then the difference between the loss under the constraint of absence and the loss
under the constraint of presence. We call this the {\em ASP confidence score}
and use this as our scoring function for both ASP variants.

Building on those two score definitions, we can now define the two metrics that
we will base our evaluation on. The first is the {\em area under the ROC
curve}, AUC ROC, given by those scores, and the second is the {\em accuracy} of
the binary classification obtained by defining a threshold for those scores: if
the score of a feature is below this threshold, the feature is considered
absent, otherwise, present. While we compute the accuracy for each SCM
separately, we compute the AUC ROC over the combined data of all the 150 SCMs.

The full evaluation procedure works as follows: We randomly choose $150$ SCMs
as described above and use them to create datasets of observations that are
then used as input for the four models. The edges and bidirected edges
estimated by the models are then compared with those of the original SCMs.
Those datasets are created in different ways, varying in the size of the
dataset and the structure of the applied interventions.

More precisely, we consider 21 different experimental setups that are applied
to the random SCMs. Those 21 setups consist of five groups:
\begin{table}[h!]
\centering
\caption{The experimental structure of the evaluation}
\begin{tabular}{ p{.5cm}|p{0.6cm}|p{6cm} } 
Int. Size & Setup ID & Intervention Sets \\
\hline
    0 & 0 & []\\
\hline
    \multirow{5}{1em}{1} & 11 & [], [0] \\
    & 12 & [], [0], [1] \\
    & 13 & [], [0], [1], [2] \\
    & 14 & [], [0], [1], [2], [3] \\
    & 15 & [], [0], [1], [2], [3], [4] \\
\hline
    \multirow{5}{1em}{2} & 21 & [], [0,1] \\
    & 22 & [], [0,1], [1,2] \\
    & 23 & [], [0,1], [1,2], [2,3] \\
    & 24 & [], [0,1], [1,2], [2,3], [3,4] \\
    & 25 & [], [0,1], [1,2], [2,3], [3,4], [4,0] \\
\hline
    \multirow{5}{1em}{3} & 31 & [], [0,1,2] \\
        & 32 & [], [0,1,2], [1,2,3] \\
        & 33 & [], [0,1,2], [1,2,3], [2,3,4] \\
        & 34 & [], [0,1,2], [1,2,3], [2,3,4], [3,4,0] \\
        & 35 & [], [0,1,2], [1,2,3], [2,3,4], [3,4,0], [4,0,1] \\
\hline
    \multirow{5}{1em}{4} & 41 & [], [0,1,2,3] \\
        & 42 & [], [0,1,2,3], [1,2,3,4] \\
        & 43 & [], [0,1,2,3], [1,2,3,4], [2,3,4,0] \\
        & 44 & [], [0,1,2,3], [1,2,3,4], [2,3,4,0], [3,4,0,1] \\
        & 45 & [], [0,1,2,3], [1,2,3,4], [2,3,4,0], [3,4,0,1], [4,0,1,2] 
\label{tableExps}
\end{tabular}
\end{table}
The first row of Table \ref{tableExps} shows the first group, consisting of a
single setup containing only the purely observational experiment, i.e. no
interventions are applied. The second row contains setups with an intervention
size equal to one. For instance, the setup with ID 11 contains two datasets
from each random SCM, the purely observational dataset (denoted by ``[]'') and
the dataset obtained from intervening on the first node (denoted by ``[1]'').
As another example, the setup with ID 15 consists of six datasets per random
SCM, the purely observed one [] and the five datasets obtained from single node
interventions on all the available nodes: intervention only on node
1 denoted by [1], intervention only on node 2 denoted by [2], and similarly for
the other nodes. In this setup 15 the input for the four causal discovery
algorithms consists of the union of those six datasets. The setups of size two
have a similar structure, except that the intervention sets now have size two.
For example, the setup with ID 23 creates as input to the four algorithms the
union of four datasets, which consists of the measurements of the purely
observational experiment and those of three (overlapping) interventions each of
size two. And finally, setups in the last row of Table \ref{tableExps} contain
the experiments of size four. Thus, each experimental setup can be described
by the combination of the size of each applied intervention and the number of
such interventions used. This enters the setup ID, where the first digit is the
intervention size and the second digit is the intervention count.

Note, that we use the same size $n$ of the dataset per intervention setting.
Thus, since we use in each experimental setup the {\em union} of the datasets
of each intervention, we get different data sizes for different setups. If the
size $n$ of a dataset is e.g.\ 1000, the setup with ID 0 will create a dataset
of size 1000, while 15 will create one of size 6000. In summary, we construct
$16$ datasets from $150$ randomly generated SCMs each, i.e.\ there are in total
2400 datasets for a given $n$. Each of our four models in the study will be
evaluated on those datasets.

We also consider different sizes $n$ of datasets. We investigate sizes 1000,
10,000, 100,000, and infinite. Here, the word ``infinite'' does not refer to
actually infinite amounts of data, but rather a version of the experiments that
correspond to its asymptotic behavior when $n\to\infty$. More precisely, this
means the following (see also \cite{llc}): When creating random SCMs, the
matrices $\BB$ and $\SSigma_\ee$ are sampled as described above. Then, we can
obtain the covariance matrix $\CC_\xx$ of the observations as follows, similar
to (\ref{cofVar}):
\begin{equation}\label{xVar}
    \CC_\xx^0 = (\II-\BB)^{-1}\SSigma_\ee(\II - \BB)^{-T}.
\end{equation}
To obtain datasets with finite size $n$, we then sample from a Gaussian
distribution with mean $\mathbf 0$ and covariance $\CC_\xx^0$, similarly for
non-zero interventions. {\em LLC} then estimates the
covariance matrix from the data and uses the estimated covariances as total
causal effects. But if we instead use the covariances in (\ref{xVar}) as the
total causal effects directly, without the detour of sampling and covariance
estimation, we obtain the total effects that we would get for an infinite amount
of data. That is meant by ``infinite'' data size for the {\em LLC} models. For
the {\em ASP} methods the approach is different: These models use the data to
obtain conditional independences. This is done by classical hypothesis tests,
see Section \ref{secAspd}. The weights in (\ref{loss}) can become infinite for
independences but stay finite for dependences for $n\to\infty$. To avoid those
asymmetries, for the infinite scenario we skip the independence tests and
collect a complete system of correct conditional dependence and independence
relations from the ground truth (we know the SCM that generated the data) into
the set $\KK$, which is thus free of contradictions, and assign to each
$k\in\KK$ the weight one, $w(k) = 1$. So this construction does not exactly
behave like {\em ASP} for $n\to\infty$, but it corresponds to it in so far as
the set $\KK$ is free of contradictions, which is why we will use the word
``infinite'' for brevity.

\begin{figure}[t]
    \centering
    \includegraphics[width=\columnwidth]{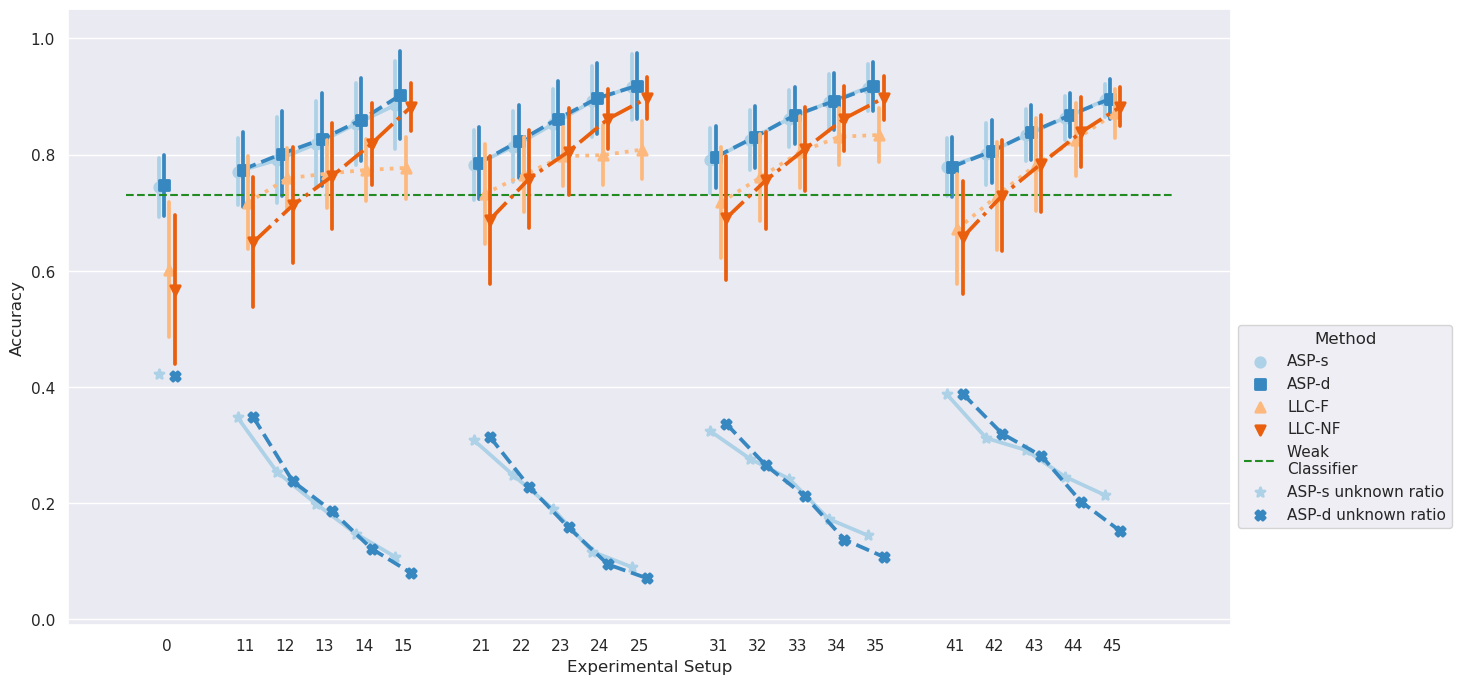}
    \caption{Accuracies sample size 1000}\label{figAcc1t}
\end{figure}
\begin{figure}[t]
    \centering
    \includegraphics[width=\columnwidth]{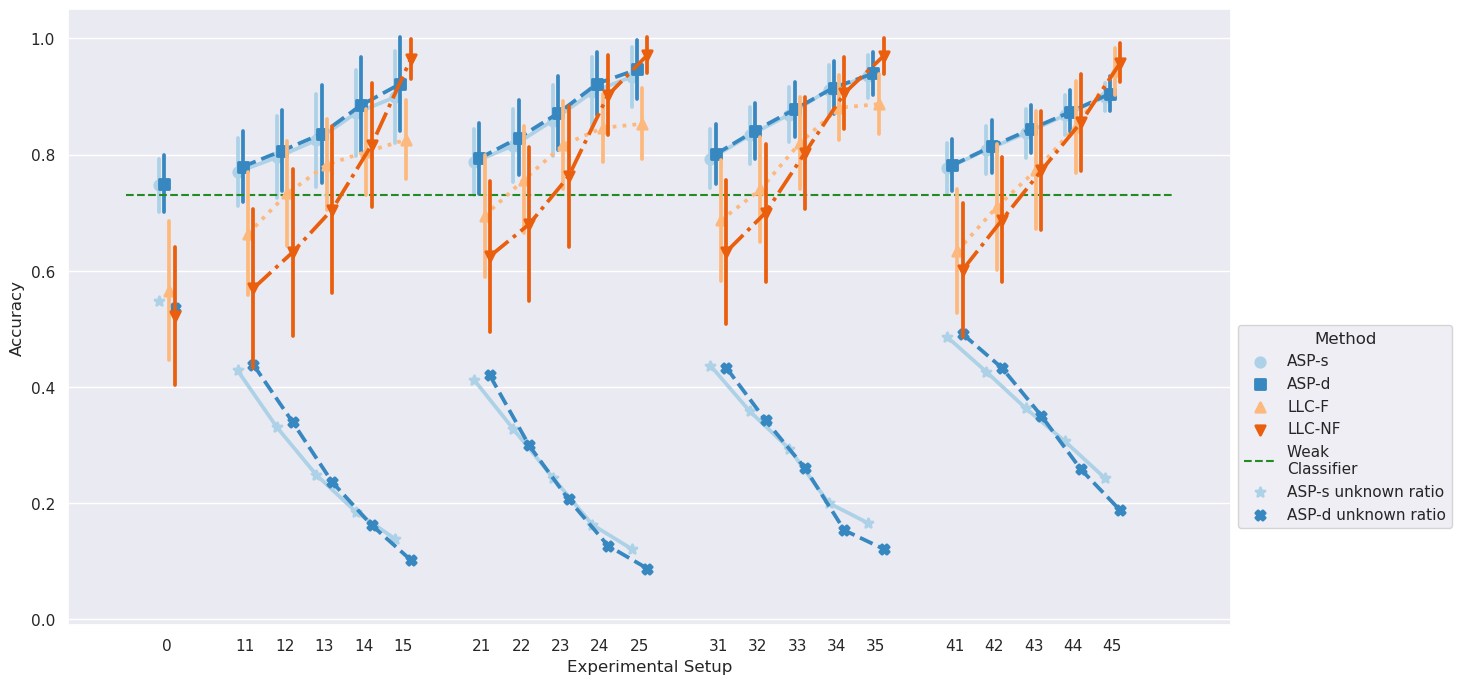}
    \caption{Accuracies sample size 10,000}\label{figAcc10t}
\end{figure}
\begin{figure}[t]
    \centering
    \includegraphics[width=\columnwidth]{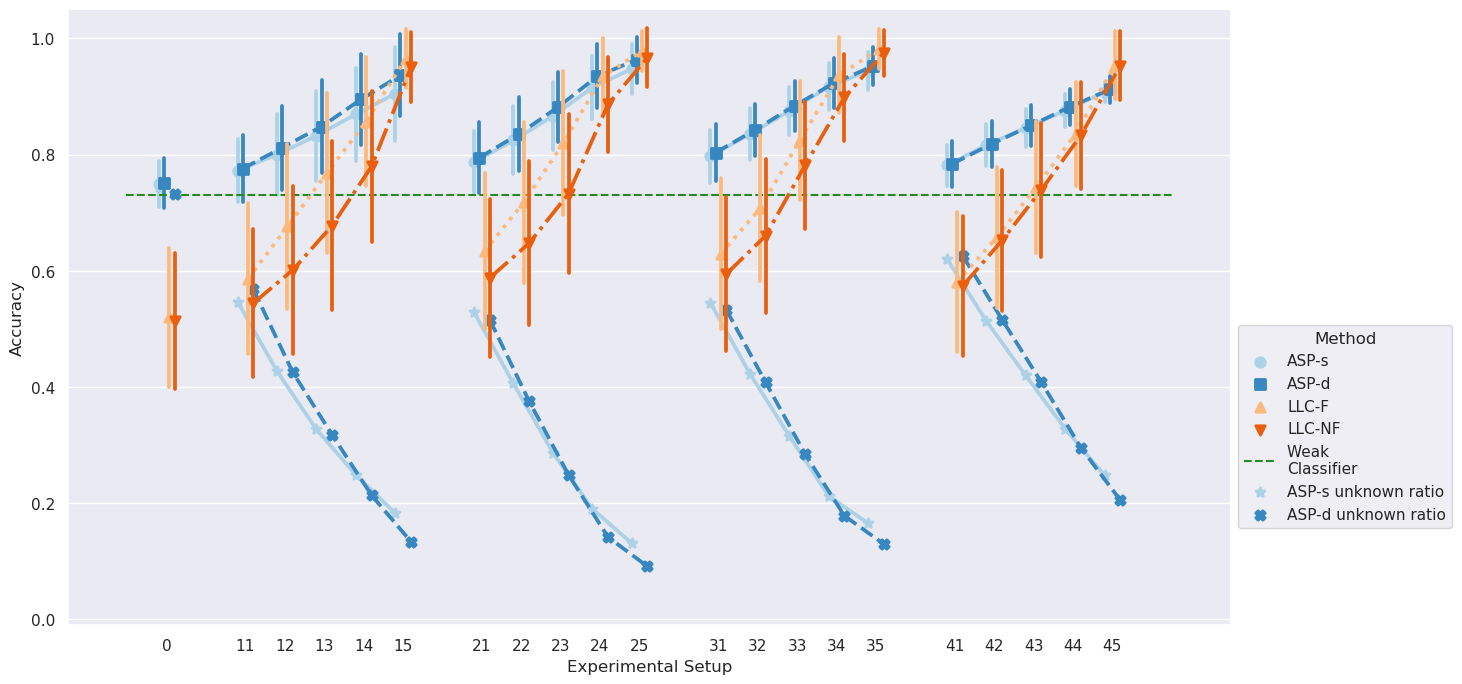}
    \caption{Accuracies infinite sample size}\label{figAccInf}
\end{figure}
\begin{figure}[t]
    \centering
    \includegraphics[width=\columnwidth]{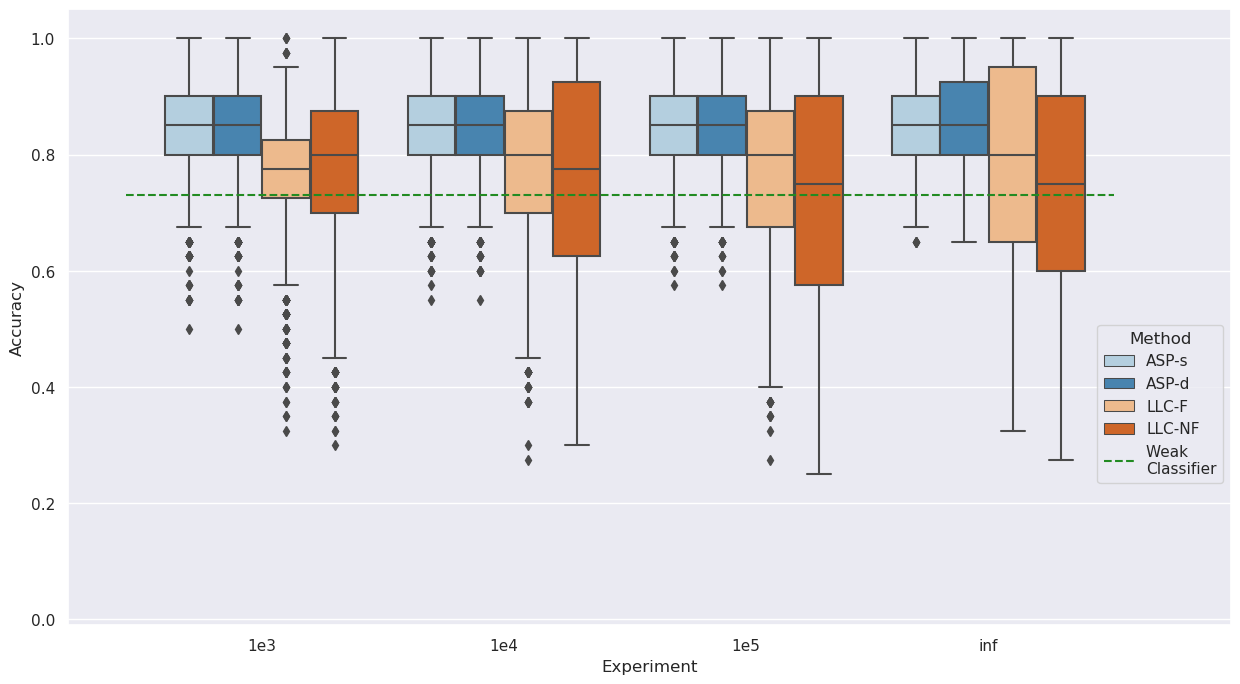}
    \caption{Average accuracies by dataset size}\label{figAccBySize}
\end{figure}

The methods we study have certain parameters that need to be fixed. For {\em
LLC-NF} it is the penalty type, $L_1$ or $L_2$, and the belonging
regularization parameter $\lambda$. For {\em LLC-F} we also have to fix the
significance level $\alpha_{LLC}$ of the conditional independence tests.
Furthermore, when using accuracy as metric, we need to choose a threshold
$t_{LLC}$ for the scores. For {\em ASP} we have the significance level
$\alpha_{ASP}$ for the conditional independence tests and, in the case of using
accuracy as a metric, again a threshold $t_{ASP}$ for the score. As with all
unsupervised learning tasks, the proper choice of those hyperparameters is
tricky in practical applications. Here, we don't consider the problem of
finding optimal hyperparameters but are rather interested in the performance
of the studied techniques if the hyperparameters are chosen roughly
appropriately, however, this is achieved. Since in our simulations we have
access to the ground truth, we can optimize the hyperparameters. For this, we
conducted rather comprehensive hyperparameter optimization studies for the type
of SCMs that get generated by our adopted sampling scheme which produces
sparse, small, cyclic models with hidden confounders as described above. This
was done using the optimization tool {\em Optuna} \cite{optuna_2019}. As usual,
we used one set of models for hyperparameter optimization and a different one
for evaluation. Based on these optimization studies we chose the following
hyperparameters for the {\em LLC} methods: penalty type $L_1$, $\lambda =
0.05$, $\alpha_{LLC} = 0.05$, and $t_{LLC} = 5$. For {\em ASP} we used
$\alpha_{ASP} = 0.05$ and $t_{ASP} = 0$.

\begin{figure}[t]
    \centering
    \includegraphics[width=\columnwidth]{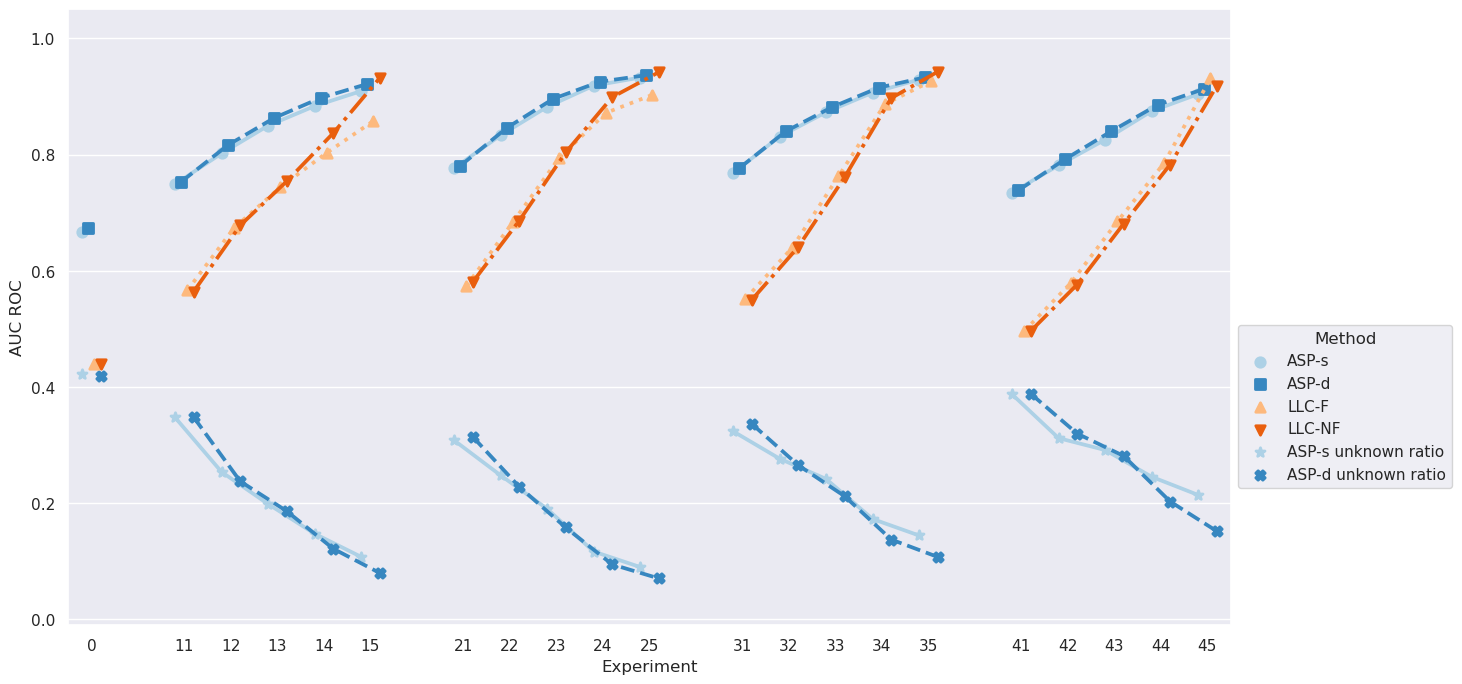}
    \caption{AUC ROC sample size 1000}\label{figAuc1t}
\end{figure}
\begin{figure}[t]
    \centering
    \includegraphics[width=\columnwidth]{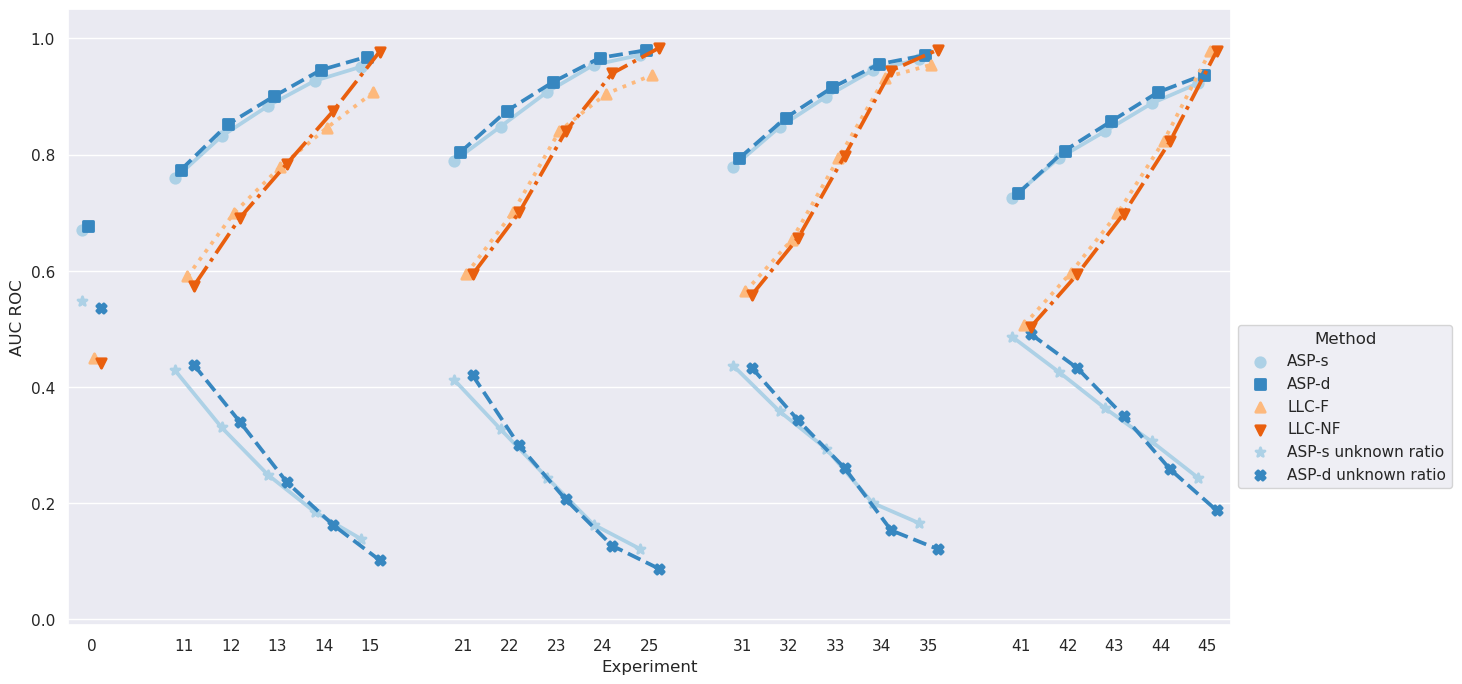}
    \caption{AUC ROC sample size 10,000}\label{figAuc10t}
\end{figure}
\begin{figure}[t]
    \centering
    \includegraphics[width=\columnwidth]{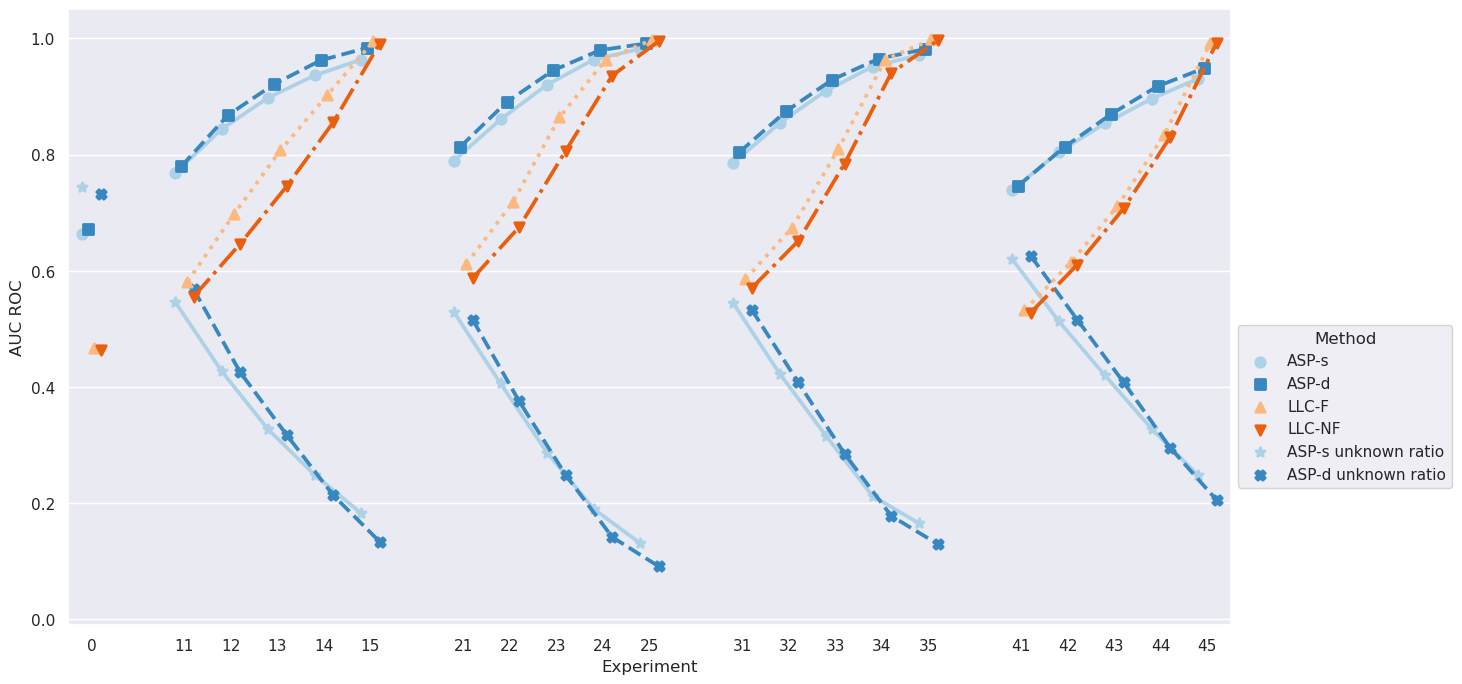}
    \caption{AUC ROC infinite sample size}\label{figAucInf}
\end{figure}
\begin{figure}[t]
    \centering
    \includegraphics[width=\columnwidth]{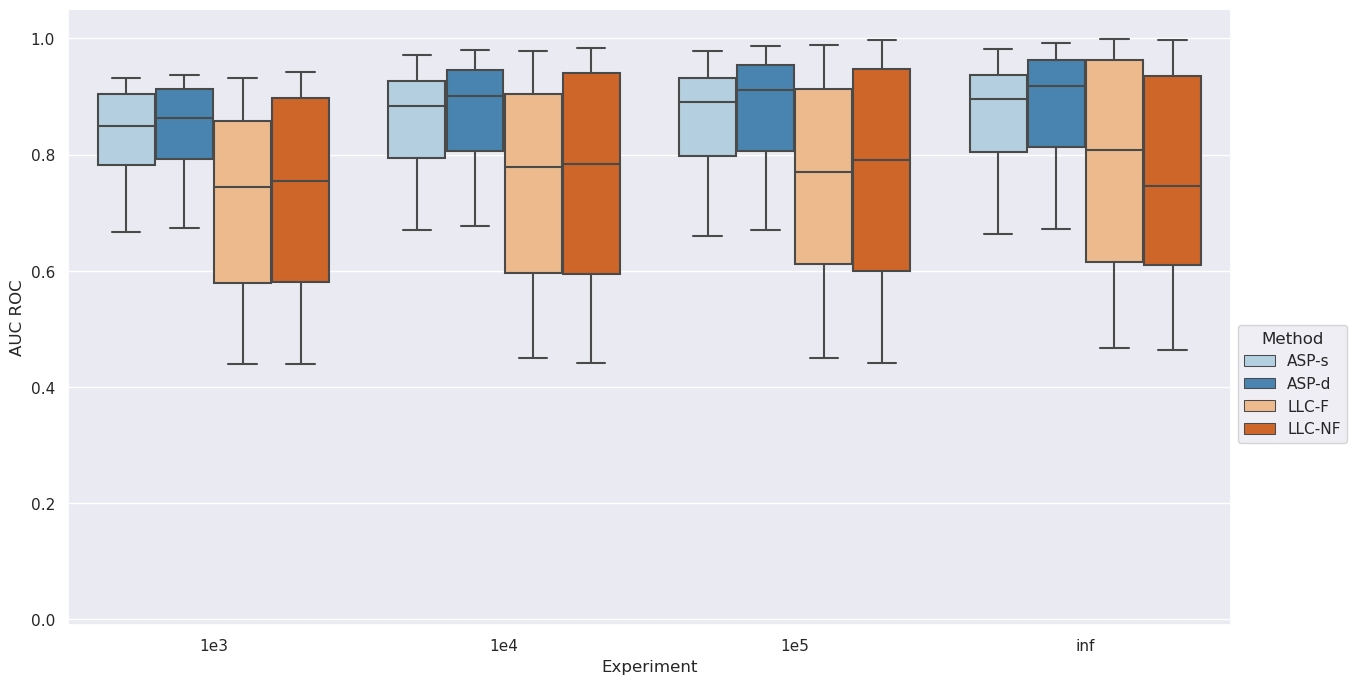}
    \caption{Average AUC ROC by dataset size}\label{figAucBySize}
\end{figure}

With those hyperparameters fixed, we ran the evaluations on the randomly
sampled SCMs with different interventional setups and multiple dataset sizes as
described above. Thus, for each combination of experimental setup, dataset
size, metric, and causal discovery method, we get 150 values for the pertaining
metric. To examine those results, we first collect the studies with accuracy as
metric and a dataset size of $n=1000$ in Figure \ref{figAcc1t}. Here, for each
experimental setup, we plot the accuracy mean together with error bars with
length equal to the standard deviation. The plots are grouped according to the
groups of the experimental setups and the four causal discovery methods are
coded by color as given in the legend. The horizontal green line is the average
accuracy of the weak classifier used in the {\em ASP}-ensembles, which just
classifies every edge as absent.

Several things can be noticed: The worst performance is measured for the purely
observational experiments, showing the relevance of interventions. Next, it is
clearly visible that combining the data of several different interventions is
beneficial in all cases, as the plots are increasing in each group. Also,
{\em LLC} requires a considerable amount of intervention data to beat the weak
classifier.
Furthermore, there seem to be no large deviations between the different sizes
of the intervenion sets, since the differences between the four large groups
are small compared to the size of the error bars. There is a slight decline in
the last groups, though, which can be explained by the fact that intervening on
all except one node removes all confounding effects, thus confounders cannot be
detected.

Maybe most importantly, the performance of the {\em ASP} methods (recall that
we use here the ensemble with the weak, sparsity-presuming classifier) seems
better than that of the {\em LLC} methods, but for the setups with five
experiments, this difference is well within the error margin. Note, that the
five-experiment setups all provide sufficient interventions to satisfy the pair
condition. It is interesting that there is almost no difference between {\em
ASP-d} and {\em ASP-s}, even though we made sure that there is at least one
cycle in each SCM. This encourages the universal use of {\em ASP-s} even if it
is not clear whether the underlying SCM is linear. The difference between {\em
LLC-F} and {\em LLC-NF} seems to switch signs within each group. This effect is
quite consistent, but its origin is not entirely clear to us, so we leave this
to future research.

In Figure \ref{figAcc10t} for dataset size $n=10,000$, we see roughly the same
behavior, except that, compared to $n=1000$, {\em LLC} seems to be worse for
smaller experiment counts and better for larger ones, beating {\em ASP} for
setups that satisfy the pair condition. The larger dataset size seems to have
no effect on {\em ASP} performance.

There is not much difference when changing to $n=\infty$, except that the
switch between {\em LLC-F} and {\em LLC-NF} within a group is not visible
anymore, and {\em LLC-F} improves considerably compared to {\em LLC-NF}. This
suggests that this switching effect for finite $n$ is due to incorrect
conditional independence tests for finite data.
As the last plot for the accuracy metric, we have averaged the performance over
all experimental setups and compared it as a function of the dataset size in
Figure \ref{figAccBySize}. Here, we see that, overall, the {\em ASP} methods have
a larger median accuracy and smaller interquartile range.

In figures \ref{figAuc1t} to \ref{figAucBySize} we visualize the performance
with respect to the total AUC ROC metric. Since it is the AUC ROC of the
combination of all the 150 SCMs, we have in each situation only a single value
which is why there is no error bar as with the accuracy metric. The conclusion
from those plots is similar to that for the accuracy metric. There is not much
difference between {\em ASP-d} and {\em ASP-s}, despite the presence of cycles.
Also, again {\em ASP} is better than {\em LLC} for fewer experiments but when
the pair condition is satisfied, {\em LLC} can beat {\em ASP}. The switching
between {\em LLC-F} and {\em LLC-NF} for finite $n$ is not very prominent,
although {\em LLC-F} gets worse for higher intervention counts. Furthermore, a
larger dataset size improves the metric. Finally, the plot in Figure
{\ref{figAucBySize} again shows higher medians and smaller interquartile range
for {\em ASP}.

\section{Conclusion}
We have evaluated four causal discovery methods that allow for cycles and
hidden confounders: two {\em LLC}-based variants {\em LLC-NF} and {\em LLC-F},
and two simple ensembles based on {\em ASP-d} and {\em ASP-s}. The study
focused on sparse linear SCMs with only five nodes and two confounders since
the {\em LLC} variants presume linearity and the {\em ASP} variants do not
scale well with the number of nodes. We considered the dependence of the
model's performance on various interventional setups and the size of the
dataset and measured them with the metrics accuracy and AUC ROC. All models
show very good discovery capabilities when applied to datasets with a
sufficient amount of interventions.  There is not much difference between {\em
ASP-d} and {\em ASP-s} even though each sampled SCM contains at least one
cycle. {\em LLC-F} is better than {\em LLC-NF} for datasets with fewer
interventions and often worse for datasets with more interventions. For
datasets with an insufficient amount of interventions, mainly thanks to the
weak classifier in the ensemble, {\em ASP} is better than {\em LLC}.

\printbibliography

\end{document}